\documentclass[10pt]{article}
\usepackage[letterpaper]{geometry}
\usepackage{hicss}
\usepackage{times}
\usepackage[none]{hyphenat}
\usepackage{url}
\usepackage{latexsym}
\usepackage[frozencache=true,cachedir=minted-cache]{minted} 
\usepackage{indentfirst}
\usepackage{graphicx}
\graphicspath{{images/}}

\usepackage{booktabs}

\title{Classifying Cyber-Risky Clinical Notes by Employing \\ Natural Language Processing}

\author{Suzanna Schmeelk$^1$, Martins Samuel Dogo$^2$, Yifan Peng$^3$, and Braja Gopal Patra$^3$\\
$^1$St. John's University,  Queens, New York\\
$^2$Queen's University, Belfast, United Kingdom \\
$^3$Department of Population Health Sciences, Weill Cornell Medicine, New York, NY, USA\\
{\underline{schmeels@stjohns.edu}}, {\underline{mdogo01@qub.ac.uk}}, {\underline{yip4002@med.cornell.edu}}, {\underline{bgp4001@med.cornell.edu}}
}

\date{}

\begin{document}
\maketitle
\begin{abstract}

Clinical notes, which can be embedded into electronic medical records, document patient care delivery and summarize interactions between healthcare providers and patients. These clinical notes directly inform patient care and can also indirectly inform research and quality/safety metrics, among other indirect metrics. Recently, some states within the United States of America require patients to have open access to their clinical notes to improve the exchange of patient information for patient care. Thus, developing methods to assess the cyber risks of clinical notes before sharing and exchanging data is critical. While existing natural language processing techniques are geared to de-identify clinical notes, to the best of our knowledge, few have focused on classifying sensitive-information risk, which is a fundamental step toward developing effective, widespread protection of patient health information. To bridge this gap, this research investigates methods for identifying security/privacy risks within clinical notes. The classification either can be used upstream to identify areas within notes that likely contain sensitive information or downstream to improve the identification of clinical notes that have not been entirely de-identified. We develop several models using unigram and word2vec features with different classifiers to categorize sentence risk. Experiments on i2b2 de-identification dataset show that the SVM classifier using word2vec features obtained a maximum F1-score of 0.792. Future research involves articulation and differentiation of risk in terms of different global regulatory requirements.

\end{abstract}

\section{Introduction}

Medical data can be misused in many ways, from business profiteering and mandatory governmental controls to an individual's identity theft. Due to the enormous impacts of identity theft and other unauthorized usages of medical data, governments are gradually developing privacy and security regulations. Examples of such regulations include the United States' Health Insurance Portability and Accountability Act (HIPAA) of 1996~\cite{usahippa}, The Health Information Technology for Economic and Clinical Health Act (HITECH) of 2009~\cite{hitech}, the European Union's General Data Protection Regulation (GDPR)~\cite{voigt2017eu}, and the Chinese Cybersecurity Law~\cite{zhang2018socio}. In addition to regulations protecting the privacy and security of clinical data, states are conforming to the United State's 21st Century Cures Act~\cite{hhs2020}, requiring healthcare providers to provide clinical data information. As healthcare data becomes more freely available, data providers need to identify security and privacy risks within shared Electronic Health Records (EHR) and to lower privacy and security risks to patients and healthcare entities. 

We classify clinical note sentences in terms of risk of containing patient health information (PHI). We solve a novel problem as prior to our contribution, researchers reported solely on methodologies for clinical note de-identification. We discuss de-identification in the literature review as we could not find any clinical note PHI risk classification literature. Our classification methodology is closely related to de-identification but unique in that classification can be employed for other purposes. For example, classification could transpire upstream or downstream from de-identification to classify risk type pre- or post- de-identification. If we implement risk classification pre- or upstream from de-identification, then we could perhaps classify different types of risks to be passed to different de-identification methodologies. If we implement risk classification post- or downstream from de-identification, then we could perhaps catch sentences that were not properly de-identified. In addition, we predict that classification may simplify the computing resources needed to keep pace with the large quantity of data being produced daily by healthcare facilities as sharing clinical notes directly with patients and for research, while maintaining privacy and security, is an international challenge. Recently, for example, \emph{OpenNotes} became an ``international movement to create partnerships toward better health and health care by giving everyone on the medical team, including the patient, access to the same information~\cite{opennotes}.''

We employ natural language processing (NLP) to identify the risk of containing protected PHI in clinical notes by classifying sentences in terms of risk based on the presence of sensitive information such as PHI. We use the Harvard Clinical NLP dataset for our analysis. The datasets were initially created at a former National Institute of Health (NIH)-funded National Center for Biomedical Computing, known as Informatics for Integrating Biology and the Bedside (i2b2)~\cite{stubbs2015annotating,kumar2015creation}. We present current literature, data, and our novel risk-classification based methodology. Finally, we present the results where the top-performing system obtained an F1-score of 0.792 for correctly identifying the risk using word2vec and SVM classifier.

The rest of the paper is organized as follows. We first present related work in Section~\ref{sec:relatedworks}. Then, we describe the data and methods in Section~\ref{sec:methods}, followed by our experimental setup, results, and discussion in Section~\ref{sec:results}. We conclude with future work in the last section.

\section{Related Work}
\label{sec:relatedworks}

NLP for patient information de-identification and risk mitigation has been developed extensively over the past decade. However, direct research for clinical note classification as introduced by our research does not exist. Therefore, to provide a state-of-the art literature review we built it on entirely different but closely related privacy and security challenges. These challenges revolve around three other pillars of literature related to de-identifying medical notes. First, there has been a need to develop accessible medical data benchmarks to evaluate de-identification methods. Second, techniques have been produced for de-identifying English language medical records using NLP. Third, the industry has expanded de-identification English-based NLP techniques to de-identify non-English medical records. Finally, to reiterate, these three pillars of de-identification research literature solve entirely different privacy and security challenges than our particular research contribution.

\textbf{Gold Standard PHI Benchmarks:} Building gold standard PHI de-identification benchmarks for NLP tasks has been an ongoing effort over the past two decades. In the early 2000s, most corpora developed to measure the performance of de-identification systems were either not publicly available for privacy reasons, or were incomplete as they were either synthetically generated or composed of only a few document types (e.g., discharge summaries, pathology reports, nursing progress notes, outpatient follow-up notes, and medical message boards). Mayer et al.~\cite{mayer2009inductive} introduced a human/manual inductive creation of an annotation schema and subsequent reference standard for de-identification of United States Veterans Affairs (VA) electronic medical records. The benchmark was created for private use only within the national VA network. The researchers manually marked PHI as risky based on the type of identifier. Deleger et al.~\cite{deleger2012building} introduced a public gold standard benchmark based on annotating different types of clinical narrative texts from the Cincinnati Children's Hospital Medical Center. Kumar et al.~\cite{kumar2015creation} introduced a new longitudinal corpus of clinical narratives based on shared task corpus from the 2014 Informatics for Integrating Biology and the Bedside (i2b2) foundation, and University of Texas Health Science Center (UTHealth) at Houston. This corpus consisted of discharge summaries and medical correspondences of 1.3K medical records for 296 patients. Further, Kumar et al.~\cite{kumar2015creation} developed a new annotated de-identification benchmark from the i2b2/UTHealth corpus. The developed de-identified datasets are limited, which affected the system performances~\cite{kumar2015creation,mayer2009inductive,deleger2012building}.

\begin{figure*}
    \centering
    \includegraphics[width=0.9\textwidth]{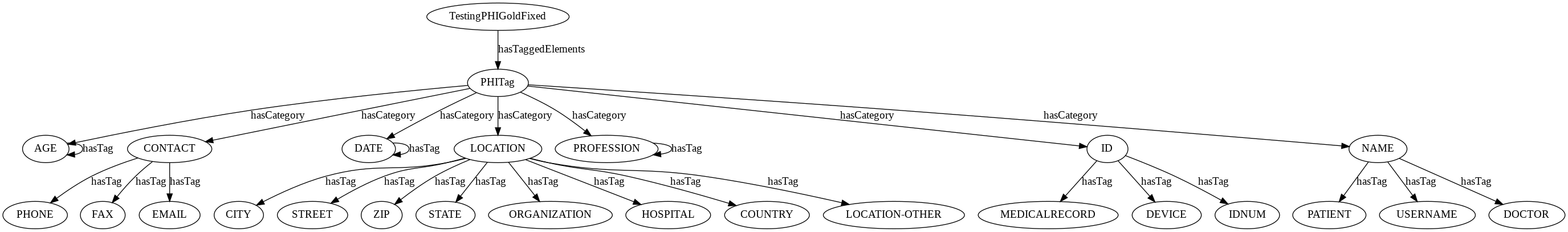}
    \caption{i2b2 PHI tag knowledge-graph}
    \label{fig:phiKnowledgeGraph}
\end{figure*}

\textbf{English De-identification Methods:} Next, we describe research on NLP systems developed on English PHI healthcare records for de-identification. In 2007, Uzuner et al.~\cite{uzuner2007evaluating} facilitated an analysis of submissions into the i2b2 project which was organized to research automating PHI removal from medical discharge records. The tool evaluation consisted of precision, recall, and F-measures on the ground truth. The systems with the best performance scored above 98\% in F-measure for all PHI categories. A few years later, Hanauer et al.~\cite{hanauer2013bootstrapping} examined the MITRE Identification Scrubber Toolkit (MIST) for record de-identification. After manual training rounds on a clinical social work and history/physical medical corpus of 360 documents, the tool improved performance based on interactive rounds. After eight hours of annotation time (round 21), MIST achieved an F1-score of 0.95. Meystre et al.~\cite{meystre2014text} extended the MIST analysis by evaluating five de-identification state-of-the-art systems: MIT, MIST (Mitre), HIDE, HMS, and MeDS on two corpora: the 2010 i2b2 NLP challenge corpus and a corpus of VA clinical notes. Overall, the authors found that the different tools had certain strengths/limitations. Liu et al.~\cite{liu2017entity,liu2017identification} performed de-identification from unstructured clinical texts via recurrent neural networks (RNNs). Liu et al.~\cite{liu2017entity} built and analyzed a long-short term memory (LSTM) and RNN model on 2010, 2012, and 2014 i2b2 challenges. Liu et al.~\cite{liu2017identification} extended the systems and analyzed the new system based on a corpus from the 2016 Centers of Excellence in Genomic Science (CEGS) Neuropsychiatric Genome-scale and RDOC Individualized Domains (N-GRID) clinical NLP challenge comprised of 1,000 (600 train and 400 test) annotated mental health records. The authors report competitive performance metrics, with limitations handling abbreviations, and specific PHI categories.

In general, the current research reflects on limitations. First, models may be built with only a small number of representative document types limiting analysis precision on document types not represented during training. Second, system evaluations trained from a certain institution document structure may not perform well on other document structures not-represented during training. Third, some parts of clinical notes, such as the medication section, may have higher error-rates than other sections of the same clinical notes. Fourth, reported models indicate limitations to how entity recognition was performed. Lastly, benchmark datasets used for evaluation can be further improved by evaluating multi-rater interrater reliability interclass correlation coefficients (ICCs)~\cite{c22224bd009f4a2a8799926e37aa789f} for more than one annotator to improve tag accuracy.

\textbf{Non-English De-identification Methods:} Non-English NLP healthcare records PHI de-identification has emerged within published research during the last few years. Jian et al.~\cite{jian2017cascaded} published research on developing a private intra-organizational Chinese de-identification benchmark, comprising 3K+ heterogeneous clinical documents, for unstructured narrative data. They reported the challenges and limitations of such English language-based techniques: (1) PHI is sparse in Chinese medical records, and (2) word segmentation and word morphological features are more difficult in Chinese than in other languages. For example, word capitalization is nonexistent in Chinese. Foufi et al.~\cite{foufi2017identification} introduced a de-identification of French unstructured clinical narrative data. Their techniques applied a Named Entity Recognition (NER) model to 11K+ French discharge summaries. They reported a limitation in the analyzed discharged summaries as they are often written in a hurry and contain as a consequence, spelling, orthographic and typographic errors, which can affect the de-identification process.

\section{Data and Methods}
\label{sec:methods}

NLP-centric risk identification appears to remain a needed, but unmet, endeavor. In this section, we explain the risk identification problem, the dataset used in our experiments, and describe the methodology that we used to identify risk in clinical notes.

\subsection{Problem}

The goal of this research is to classify clinical note sentences into two risk categories, high or low, based on whether or not each sentence is at risk of potentially leaking sensitive information about patients. The input is a sentence in a medical note. The output is one of the two classes of risks depending on the presence of sensitive words within the sentence.

\subsection{i2b2 De-identification Data}

In this project, we used the i2b2 de-identification dataset to evaluate the proposed models~\cite{kumar2015creation}. We used BeautifulSoup~\cite{richardson2007beautiful} to parse each clinical note into sections for analysis of the actual text and analysis of the gold-standard PHI tags. 

\begin{figure*}[!htbp]
    \centering
    \includegraphics[width=0.9\textwidth]{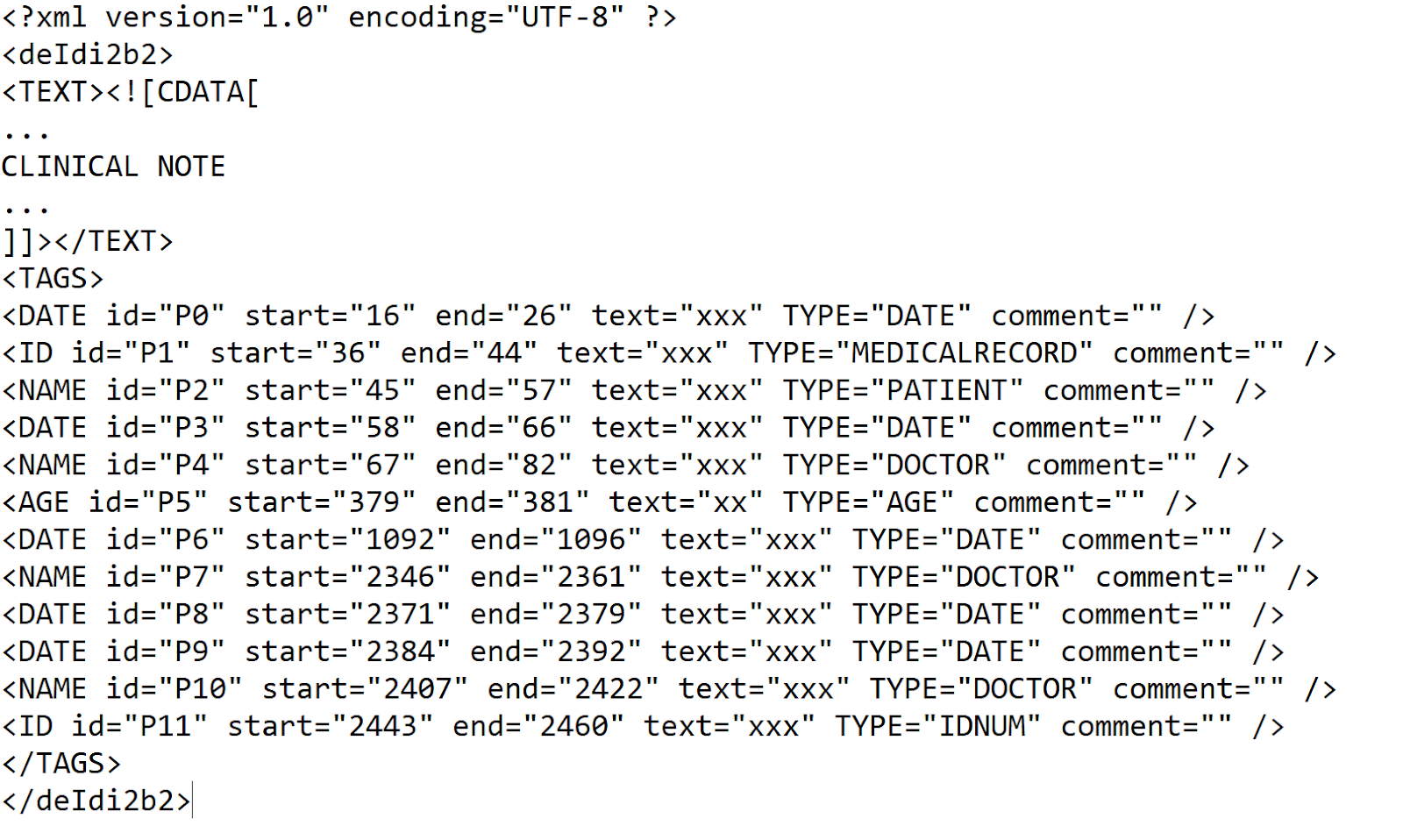}
    \caption{A sample clinical note structure in the i2b2 de-identification dataset}
    \label{fig:data}
\end{figure*}

Each gold-standard dataset file consists of an XML file with a section for the actual clinical note wrapped as a $<$TEXT$><$![CDATA[$\dots$]]]$><$/TEXT$>$ and a separate section $<$TAGS$><$/TAGS$>$ listing the gold standard sensitive information locations within the note. For example, tags include gold standard identifiers and locations within the clinical notes for sensitive PHI information such as specifics related to HOSPITAL, DATE, DOCTOR, USERNAME, NAME, MEDICALRECORD, AGE, and IDNUM. Figure~\ref{fig:phiKnowledgeGraph} shows our created knowledge-graph representation of the PHI categories and tags within the analyzed i2b2 dataset. This dataset has labeled PHI into eight main categories: Age, Contact, Date, Location, Profession, ID and Name. Figure~\ref{fig:data} shows the layout of a gold-standard PHI labeled clinical note. All sensitive information has been removed in the figure. As shown in the XML markups, each tag provides a location for the sensitive information within the clinical note.

During the data cleaning, we split the notes into sentences, using the Natural Language Toolkit (NLTK)~\cite{bird2009natural}, and rearranged tags to the sentences. Each sentence was then given a risk ranking based on the gold-standard PHI labels. In fact, sentences may contain more than one tag and could further be classified on a larger risk scale in future research, but in this project, we only focus on binary classes: low (0) or high (1). In total, 22,541 sentences were given risk scores from the tagged 514 clinical note dataset (Table~\ref{data}). We used five folds to group the dataset into training and testing. The characteristics of the dataset in terms of the number of sentences with tags (high risk) and the number of sentences without tags (low risk) is seen in Table~\ref{data}.

\begin{table}[]
\centering
\begin{tabular}{lr}
\toprule
&  \textbf{Total} \\ 
\midrule
\textbf{Low} & 11,750 \\ 
\textbf{High} & 10,791 \\ 
\textbf{Total} & \textbf{22,541} \\ 
\bottomrule
\end{tabular}
\caption{Characteristics for the dataset (\# of sent.)}
\label{data}
\end{table}

\subsection{Features}

\textbf{Bag-of-Words Model:} We used the NLTK~\cite{bird2009natural} to transform clinical note sentences into unigrams, or bag-of-words (BOW). After the transformation, the number of times a unigram appears in each text is counted as the feature~\cite{jurafsky}. The number of features in this model is equal to the vocabulary size found by analyzing the data. We employed scikit-learn CountVectorizer~\cite{pedregosa2011scikit} with the binary parameter set to true so that all non zero counts are set to one. Setting the binary parameter as such is useful for discrete probabilistic models that model binary events rather than integer counts.

\textbf{Word-Embedding Model:} We also employed word2vec to transform words into vectors. We used the word2vec model pre-trained on the Google News dataset which gives a 300 dimensional vector. The word vectors were averaged to form the final sentence vector using the spaCy word2vec libraries~\cite{honnibal2017spacy}. 

\textbf{Experimental Settings:} We employed classifiers with different parameters. The experimental settings and results obtained are discussed in Section~\ref{sec:results}. 

The risk classification was performed on the Harvard I2B2 N2C2 project dataset. Specifically, the 2014 De-identification \& Heart Disease dataset. Within the 2014 i2b2/UTHealth corpus we examined the tagged dataset testing-PHI-Gold-fixed which is part of the deIdi2b2 subgroup. Table~\ref{phirisk-metadata} shows the PHI labeled tag counts within the dataset that we calculated matching the tag counts reported by He et al.~\cite{pmid26315662}. 

\begin{table}[]
\centering
\begin{tabular}{lr}
\toprule
\textbf{Tag Name} & \textbf{Count}\\ 
\midrule
\textbf{EMAIL} &	1 \\ 
\textbf{FAX} & 	2 \\ 
\textbf{DEVICE} & 	8 \\ 
\textbf{LOCATION-OTHER} & 	13 \\ 
\textbf{ORGANIZATION} & 	82 \\ 
\textbf{USERNAME} & 	92 \\ 
\textbf{COUNTRY} & 	117 \\ 
\textbf{STREET} & 	136 \\ 
\textbf{ZIP} & 	140 \\ 
\textbf{PROFESSION} & 	179 \\ 
\textbf{STATE} & 	190 \\ 
\textbf{IDNUM} & 	195 \\ 
\textbf{PHONE} & 	215 \\ 
\textbf{CITY} & 	260 \\ 
\textbf{MEDICALRECORD} & 	422 \\ 
\textbf{AGE} & 	764 \\ 
\textbf{HOSPITAL} & 	875 \\ 
\textbf{PATIENT} & 	879 \\ 
\textbf{DOCTOR} & 	1912 \\ 
\textbf{DATE} & 	4980 \\ 
\midrule
\textbf{Total} & 	\textbf{11462} \\ 
\bottomrule
\end{tabular}
\caption{Characteristics for PHI tags in the dataset}
\label{phirisk-metadata}
\end{table}

We employed five classifiers from scikit-learn~\cite{pedregosa2011scikit}. First, in the first experiment for each model, we employed the Bernoulli Na\"ive Bayes (NB) Classifier, which accepts binary values, is used for discrete data with a Bernoulli distribution. Next, we employed the Gaussian NB Classifier, which is useful when working with continuous values with probabilities that can be modeled using a Gaussian distribution. Third, we employed the AdaBoost Classifier, which is a meta-estimator that adjusts outlying cases during classification such that subsequent classifications focus on difficult cases. Fourth, we employed the RandomForest Classifier, which is a meta estimator that fits a number of decision tree classifiers on various sub-samples of the dataset and uses averaging to improve the predictive accuracy and control over-fitting, with the default value of 100 trees. Fifth, we employed the C-Support Vector Machine (SVM) classifier, a statistical learning framework with an implementation based on \emph{libsvm}. Lastly, we employed LinearSVM which is computationally more efficient than standard SVM as it uses a linear kernel.

For all the classifiers, the default scikit-learn parameters were employed. For the Gaussian NB Classifiers, the prior probabilities of the classes were not specified and a portion of the largest variance of all features was used at $1\mathrm{e}{-9}$. For the Bernoulli analysis, we employed the default parameters to enable Laplace/Lidstone smoothing. We also set the threshold for binarizing to the default of none as it was presumed to already consist of binary vectors. Similarly, the AdaBoost four parameters were set to the defaults as well as RandomForest, LinearSVM, and SVM.


\subsection{Evaluation Metrics} 

We evaluated the mode using precision, recall, and F1-score. Precision is defined to be the proportion of cases classified as positive that were true positives. In our case, the precision identifies the number of high risk sentences that were in fact classified as having high risk in that they did contain sensitive patient information. Recall, or sensitivity, is defined to be the proportion of positive cases in the gold standard that are correctly classified as positive. In our case, high risk sentences that were identified as having high risk. 

\section{Results and Discussion}
\label{sec:results}

We examine performances of classifiers. We were able to achieve different results with different classifiers using Bag-of-words (BOW) features as summarized in Table~\ref{results}, which show the mean of 5-fold cross-validation for the five classifiers. The LinearSVM classifiers obtained the best cross-validated F1-score of 0.767.

Next, classifiers using word2vec feature out performed previous models using BOW. The results in Table~\ref{results} show the mean of 5-fold cross-validation for the five classifiers. The SVM classifier obtained the best cross-validated F1-score of 0.792. We will further examine the measurements in Table~\ref{results} as interpreted in our case of classifying/identifying sentences within clinical notes that contain sensitive patient information. 

\begin{table*}[!htbp]
\centering
\begin{tabular}{llrrr}
\toprule
 & Model & Precision & Recall & F1-score\\ \midrule
BOW & Bernoulli Na\"ive Bayes & 0.9062 & 0.4977 & 0.650  \\ 
 & Gaussian Na\"ive Bayes & 0.8043 & 0.5737 & 0.659 \\ 
 & AdaBoost & 0.7994 & 0.5763 & 0.652 \\ 
 &  RandomForest & 0.8670 & 0.6787 & 0.757 \\
 & LinearSVM & 0.8266 & 0.7392 & \textbf{0.767} \\
 & SVM & 0.8163 & 0.7078  & 0.758 \\\midrule
word2vec & Bernoulli Na\"ive Bayes & 0.7470 & 0.6761 & 0.717 \\
& Gaussian Na\"ive Bayes & 0.6949 & 0.7676 & 0.742 \\ 
& AdaBoost & 0.7619 & 0.7240 & 0.752 \\ 
& RandomForest & 0.8242 & 0.7228 & 0.767 \\
& LinearSVM & 0.7940 & 0.7397 & 0.765 \\
& SVM & 0.8249 &  0.7583 & \textbf{0.792} \\
\bottomrule
\end{tabular}
\caption{Model performances}
\label{results}
\end{table*}

Table~\ref{results} shows that different classifiers give different precision and recall results. In our case of identifying high-risk sentences, the recall measurement has stronger security ramifications than precision ramifications since recall measures the proper detection/classification of high risk sentences within the clinical notes. Recall takes into account the false negatives into the calculation. The F1-score compares precision measures with recall measures and in general higher F1-scores are considered stronger performance.

\begin{figure}[]
    \centering
    \includegraphics[width=0.4\textwidth]{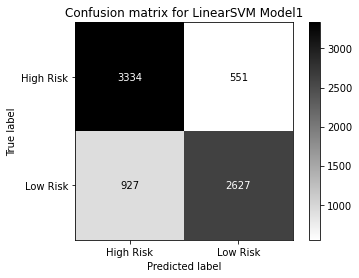}
    \caption{Confusion matrix for system using LinearSVM with BOW features}
    \label{fig:confusion1}
\end{figure}

\begin{figure}[]
    \centering
    \includegraphics[width=0.4\textwidth]{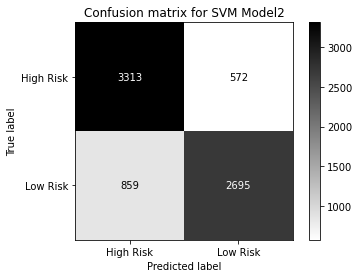}
    \caption{Confusion matrix for system using SVM with word2vec features}
    \label{fig:confusion2}
\end{figure}

Figure~\ref{fig:confusion1} (a) BOW and \ref{fig:confusion2} (b) word2vec show the confusion matrix for the top two performing classifiers as they had the overall best performance. The top-right corner is the measure of most importance for privacy and security as it indicates the number of sentences within the clinical notes which contained sensitive patient information that were labeled improperly as low risk. Figure~\ref{fig:confusion1} shows that the SVM classifier improperly labeled 551 sentences as low risk; and Figure~\ref{fig:confusion2} shows that the SVM classifier improperly labeled 572 sentences as low risk when in fact it contained sensitive patient information. 

\textbf{Error analysis:} Further, we conducted error analysis to examine why our model may make errors. There are three steps in our model which may have errors. First, the data parsing of the clinical notes using NLTK saw the best performance over other parsing libraries, but this could still be improved for more properly parsing medical texts. Second, the feature extraction could also be improved. Perhaps, our feature extraction methods may not be adequately identifying short/long sentences within the clinical notes. Third, more advanced methods could be employed such as grid search to find optimal parameters for the classifiers used in this work. These directions of future work may improve our current state-of-the-art model.

\section{Conclusion and Future Work}
\label{sec:conclusion}

The identification of risk and the de-identification of medical records is now essential to uphold privacy and security regulations of patient health information employed for medical research and information sharing. Early literature in this domain consisted of developing industry benchmarks for the evaluations of de-identification techniques and methodologies starting with private datasets moving to public gold standard datasets. With the development of benchmarks, the healthcare industry was enabled to further openly analyze the performance of de-identification techniques employing NLP. As the research showed, de-identification success with English-based medical records. With the emergence of translational services such as Google Translate, the latest industry research has been in exploring NLP de-identification techniques in other languages such as Chinese, French, Serbian, German, Korean, Portuguese and Spanish. Privacy and security concerns will only mature with time.

In this research, we identified NLP as a useful methodology to identify sentence risk of containing sensitive information within clinical notes. Our developed systems were able to achieve the state-of-the-art of such a novel need. Future work involves improving the data cleaning, sentence vectorization, classification, and the elaboration of sentence risk levels. For risk levels, we could classify sensitive information on a risk scale where identifiers such as hospital name, doctor name, and perhaps age carry less patient identifying risk than a patient name, a patient medical record number, and other direct patient identifiers. As clinical notes need to be de-identified for data sharing and research, this work provides fundamental contributions for employing NLP to enable more robust healthcare.

\section*{Acknowledgment}

This work is supported by the National Library of Medicine under Award No. 4R00LM013001.

\bibliographystyle{ieeetr}
\bibliography{sample}

\end{document}